# Fusion or Confusion? Assessing the impact of visible-thermal image fusion for automated wildlife detection

Camille Dionne-Pierre[1,2], Samuel Foucher[1], Jérôme Théau[1,2], Jérôme Lemaître[3], Patrick Charbonneau[3], Maxime Brousseau[4], Mathieu Varin[4]

[1]Department of Applied Geomatics, Université de Sherbrooke, 2500 Boulevard de l'Université, Sherbrooke, J1K 2R1, Québec, Canada

[2]Quebec Centre for Biodiversity Science, McGill University, Stewart Biology Building, Department of Biology, 1205 Dr. Penfield Avenue Montreal, H3A 1B1, Quebec, Canada

[3]Ministère de l'Environnement, de la Lutte contre les Changements Climatiques, de la Faune et des Parcs (MELCCFP), 880 Chemin Sainte Foy, G1S 4X4, Québec, Canada

[4]Centre d'enseignement et de recherche en foresterie de Sainte-Foy, 2440 chemin Sainte-Foy, G1V 1T2, Québec, Canada

**Abstract**
Efficient wildlife monitoring methods are necessary for biodiversity conservation and management. The combination of remote sensing, aerial imagery and deep learning offer promising opportunities to renew or improve existing survey methods. The complementary use of visible (VIS) and thermal infrared (TIR) imagery can add information compared to a single-source image and improve results in an automated detection context. However, the alignment and fusion process can be challenging, especially since visible and thermal images usually have different fields of view (FOV) and spatial resolutions. This research presents a case study on the great blue heron (*Ardea herodias*) to evaluate the performances of synchronous aerial VIS and TIR imagery to automatically detect individuals and nests using a YOLO11n model. Two VIS-TIR fusion methods were tested and compared: an early fusion approach and a late fusion approach, to determine if the addition of the TIR image gives any added value compared to a VIS-only model. VIS and TIR images were automatically aligned using a deep learning model. A principal component analysis fusion method was applied to VIS-TIR image pairs to form the early fusion dataset. A classification and regression tree was used to process the late fusion dataset, based on the detection from the VIS-only and TIR-only trained models. Across all classes, both late and early fusion improved the F1 score compared to the VIS-only model. For the main class, occupied nest, the late fusion improved the F1 score from 90.2 (VIS-only) to 93.0%. This model was also able to identify false positives from both sources with 90% recall. Although fusion methods seem to give better results, this approach comes with a limiting TIR FOV and alignment constraints that eliminate data. Using an aircraft-mounted very high-resolution visible sensor (wide FOV) could be an interesting compromise for operationalizing surveys.

## 1. Introduction

With the world now facing a biodiversity crisis (Clarfeld et al., 2025), effective monitoring methods are crucial to support wildlife management (Fuller et al., 2020). Conventional wildlife detection methods, such as direct observation, track surveys or camera traps, have limitations that can affect their efficiency and accuracy (Silveira et al., 2003). These methods are difficult to employ in remote or hard to access habitats and are often dependant on the expertise of the field personnel (Silveira et al., 2003). Innovative technologies can help to overcome these limitations by complementing or replacing conventional methods.

Aerial imagery represents an opportunity for wildlife monitoring, more specifically for wildlife counts (Hollings et al., 2018). Images are a form of permanent data, as they can be archived and reviewed at a later time, unlike visual surveys (Hodgson et al., 2013). Aerial images can be acquired by diverse platforms such as planes and drones to survey remote or hard to access places (Hollings et al., 2018) using different types of sensors. For wildlife detection, visible (VIS) imagery sensors are the most commonly used (Elmore et al., 2023) but thermal infrared (TIR) sensors have also been used to detect several species such as white-tailed deer (*Odocoileus virginianus*) (Beaver et al., 2020; Chrétien et al., 2016), koalas (*Phascolarctos cinereus*) (Corcoran et al., 2019), livestock (Krishnan et al., 2023) and rails (Olsen et al., 2023). However, TIR imagery is less utilised than VIS imagery due to its lower spatial resolution (Chabot and Francis, 2016; Linchant et al., 2015). While often used on their own, certain studies have combined VIS and TIR imagery for a common task, relying on the manual interpretation of the complementary information each provides (Kays et al., 2018; McKellar et al., 2021).

A large number of images are usually collected during a drone-based survey because of the low flight altitude and the compact footprint of such imagery (Chabot and Francis, 2016). This poses a problem with the time required to manually analyse and review each of these images (Chabot and Francis, 2016) and with manual counting, which is subject to observer bias (Milton et al., 2006). The development of automated wildlife detection in aerial imagery represents an alternative to the time-consuming manual analysis. Object detection methods based on convolutional neural networks (CNNs) have demonstrated a strong ability to detect animals on different types of imagery (Barbedo et al., 2019; Delplanque et al., 2023b; Eikelboom et al., 2019; Kellenberger et al., 2018; Peng et al., 2020).

Although CNNs have been frequently used on VIS imagery for animal detection, their use on TIR imagery remains limited. Automated detection of deer from the ground has been tested on thermal imagery acquired at night (Mowen et al., 2022; Munian et al., 2022; Popek et al., 2023). TIR imagery can provide information that is overlooked in VIS images, especially for cryptic animals that blend well with their environment (Krishnan et al., 2023). Combining information from VIS and TIR sensors can aid animal detection by better differentiating subjects from their surroundings and can yield better results for automated detection with CNN. In Yadav et al. (2020), a simple CNN fusion architecture was used to detect objects in an automated driving system using VIS and TIR images. The fusion method outperformed the unimodal VIS and TIR baselines on an aligned dataset, but did not perform well on a different dataset containing images with extremely different fields of view (FOVs) and resolutions, outlining a need for further research. In Krishnan et al. (2023), VIS and TIR image fusion was tested, but the limited dataset made it difficult to draw conclusions as to the added value of this technique. Furthermore, this study used older versions of YOLO models (YOLOv5 and YOLOv7). Newer YOLO models are now available, potentially offering better performances for this specific task.

Different multi-sensor data fusion methods can be used to extract information from both VIS and TIR images (Khaleghi et al., 2015). However, the different FOVs of VIS and TIR sensors poses a challenge when using data fusion (Song et al., 2024). Misaligned VIS-TIR images can make detection models less accurate, making image alignment a critical step. Existing VIS-TIR datasets are often manually aligned, which can be time-consuming (Song et al., 2024). To alleviate this problem, deep neural networks can also be used to automatically align images by finding correspondences between two images (Lindenberger et al., 2023). For data fusion approaches, a distinction is made between early and late fusion (Baltrušaitis et al., 2019). Early fusion methods fuse different images to integrate them into a new fused image (Gadzicki et al., 2020; Pereira et al., 2023) that can then be used in training to detect objects (Krishnan et al., 2023). The hypothesis is that this new image contains more information than the source images (Sun et al., 2020). In late fusion methods, the source images are used to train independent models and the results are fused at the end of the process (Pereira et al., 2023). Both methods have the potential to yield a better precision in object detection tasks. To our knowledge, there are no published studies to date that utilize late fusion or compare late and early fusion methods on VIS and TIR aerial imagery for wildlife detection.

To test and compare the applicability of early and late fusion methods for automated wildlife detection, a case study is presented on the great blue heron (*Ardea herodias* GBHE; hereinafter also named as heron), a large and common wading bird. This colonial waterbird builds large nests in trees in heronries (Short and Cooper, 1985). Heronries with ≥ 5 nests are legally protected on public land in the province of Québec (Canada) by Quebec law (C-61.1, r. 18). These heronries are surveyed every 5 years. These surveys are currently performed by helicopter and visual counting. Heronries in Quebec therefore present a good case study for testing the potential of automatic image-based detection. Such an approach could optimize surveys by reducing costs and efforts, while potentially increasing their frequency. In these surveys, the nests and the herons are counted. These two targets differ widely in terms of size, color, texture and temperature. VIS and TIR images can thus pick up different information for these targets, making them an excellent subject for testing data fusion and its added value to automated detection. This study aimed to evaluate VIS and TIR aerial image fusion for automated detection of GBHE nests and individuals comparing early and late fusion methods, and single source imagery.

## 2. Methods

### 2.1. Study area

Images were acquired from six heronries located in different administrative regions of Quebec, Canada, between June 5$^{th}$ and July 5$^{th}$ during the summer of 2023 (Table 1, Figure 1). The heronries were in either aquatic (beaver ponds) or forested environments. The total number of surveyed nests was 310, with the majority (240 nests) located in the *Grande Île* heronry.

**Table 1 :** Heronry type and number of surveyed nests for each heronry visited in the summer of 2023.

| Id | Survey date | Numbers of surveyed nests | Heronry type |
|----|-------------|---------------------------|--------------|
| 1 | June 5 | 20 | Aquatic |
| 2 | June 8 | 16 | Forest |
| 3 | June 19 | 12 | Aquatic |
| 4 | June 20 | 10 | Aquatic |

| Id | Survey date | Numbers of surveyed nests | Heronry type |
|----|-------------|---------------------------|--------------|
| 5  | June 21     | 12                        | Forest       |
| 6  | July 3-5    | 240                       | Forest       |

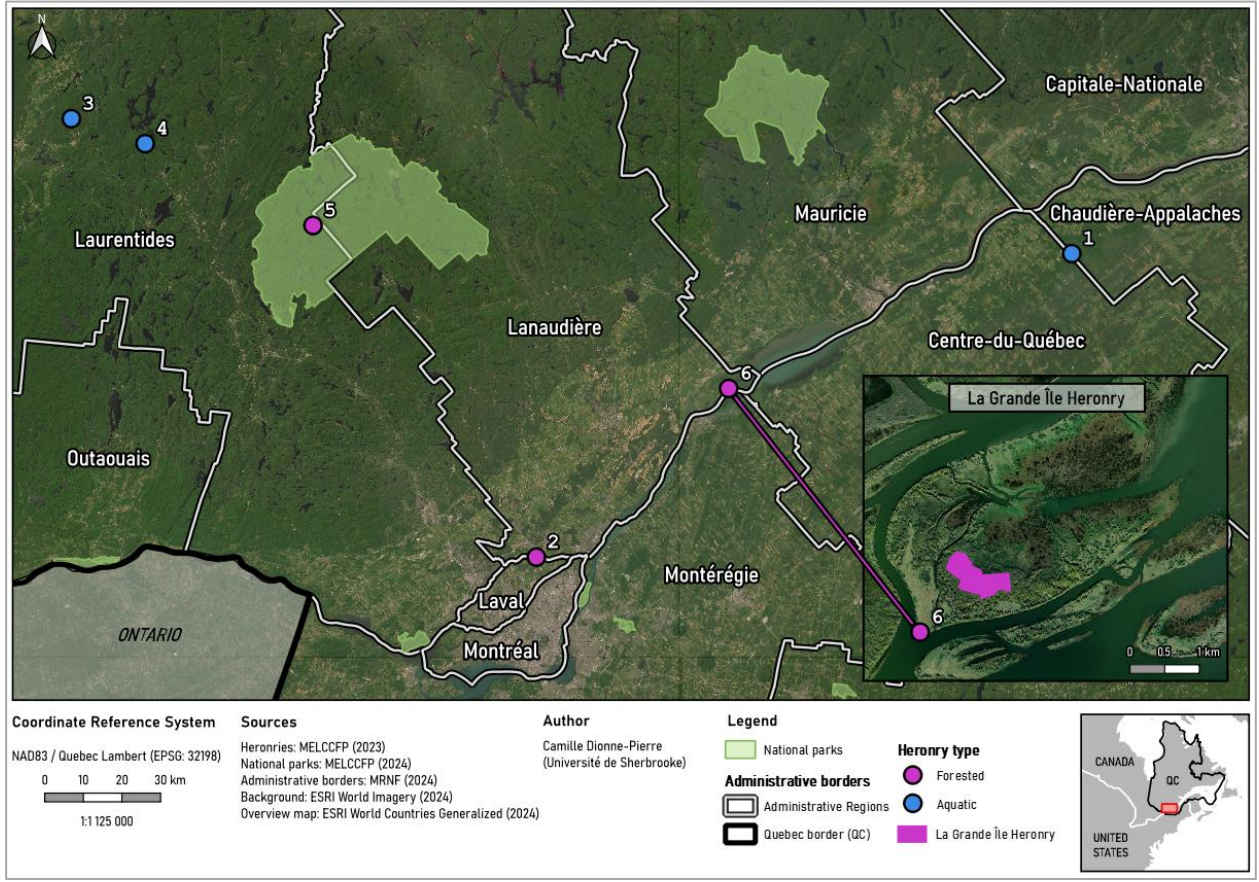

**Figure 1 :** Localisation of the heronries visited in the summer of 2023 by heronry type and Quebec administrative region.

### 2.2. Data acquisition

Drone imagery was acquired using a DJI Zenmuse H20T mounted on a DJI Matrice 300 drone (SZ DJI Technology Co., Ltd, Shenzhen, China), enabling us to simultaneously obtain a TIR image containing one band (640 × 512 pixels) and a VIS image containing a red, green and blue band (RBG) (4056 × 3040 pixels) of the same target. Each mission had five flight plans (1 nadir and 4 oblique). A total of 9932 VIS-TIR image pairs were taken (2062 nadir and 7870 oblique), with 20% of these containing GBHE nests.

All flight plans were set at a minimum altitude of 50 m above ground level, which was determined by reviewing the literature and through field tests conducted by the *ministère de l'Environnement, de la Lutte contre les changements climatiques, de la Faune et des Parcs* (MELCCFP) to minimize disturbance to GBHEs (Barr et al., 2020; Charbonneau and Lemaître, 2021; Geldart et al., 2022; Zbyryt et al., 2021). The drone pilots were all certified and all field activities were approved in advance by the MELCCFP Animal Care Committee in the form of animal care certificates (CPA-Faune 22-18, CPA-Faune 23-16).

### 2.3. Image processing

#### 2.3.1. Image alignment

Image alignment was crucial to ensure that objects of interest were superimposable and comparable (Gadzicki et al., 2020). Before being labeled, the pairs of TIR and VIS images were aligned using LightGlue (Lindenberger et al., 2023), a feature matching framework based on a deep neural network trained to recognize common local features between two images. Since the VIS image has a FOV about 5 times larger than the TIR image (Figure 2), a center crop was performed on the VIS image using a reduction factor of 0.6. LightGlue first found matching keypoints between the two images and then applied an affine transformation from the VIS image to the TIR image. Alignment was performed prior to manual annotation, as many of the image pairs could not be aligned with sufficient accuracy and therefore were not annotated (Figure 3). Image pairs with too much alignment error were removed from the dataset, firstly by using the number of keypoints (≤45) and the average squared difference between keypoints (>68 px), and secondly by a visual verification. The final aligned image size was 1792 x 1433 pixels.

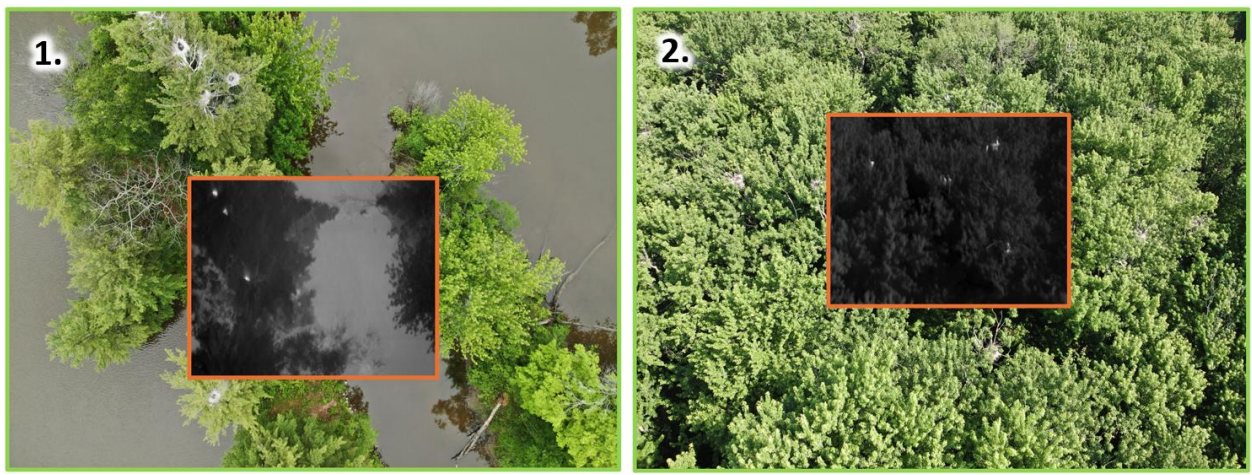

**Figure 2 :** Example of the field of view difference between corresponding visible image (green) and thermal image (orange) on a pair of nadir (1) and oblique (2) images.

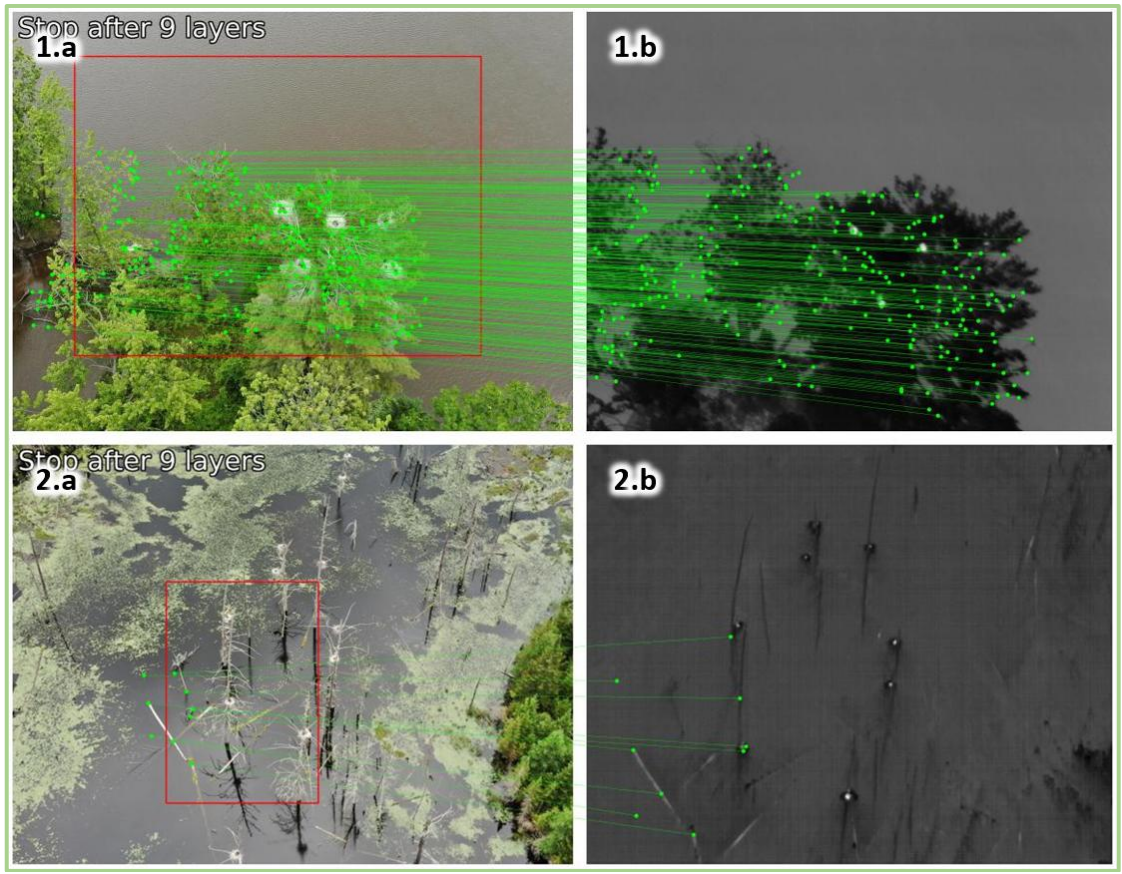

**Figure 3 :** Matching of keypoints on corresponding visible (VIS) and thermal (TIR) images. VIS image (1.a) and its corresponding TIR image (1.b) show a sufficient number of matches between keypoints (>45). VIS image (2.a) and its corresponding TIR image (2.b) show a low number of matches between keypoints (≤45). The red rectangles on the two VIS images represent the derived footprints of the aligned TIR images.

### 2.3.2. Annotation

Once aligned, images were transferred to the Segments.ai platform (Segments.ai, 2024) for labelling, using bounding boxes. The objects labelled on the VIS images were occupied nests, empty nests and isolated individuals (herons outside nests). The individuals inside the nests were not labeled since each heron was difficult to distinguish from another due to the spatial resolution of the VIS image. In the TIR images, as nests are rarely visible, the isolated individuals and the individuals inside nests were labeled in a single class named "heron". In some instances, several individuals could be very close to one another and were indistinguishable. In this case, the hot spot in the TIR image was labeled with one bounding box.

### 2.4. Automated detection with YOLO11n

The model used was YOLO11n, a real-time object detector from the YOLO series (Ultralytics, 2025); the latest version of YOLO at the time of the analysis. The nano (n) model was used, as its smaller size enables training to be carried out quickly, taking up less space on the graphics processing unit while still delivering good performance (Ultralytics, 2025). The number of training epochs was set at 50 and the images were cut into 640 × 640 pixel patches, to be compatible with the YOLO11n input layer. This model was used for two types of fusion methods: early fusion and late fusion. The models' performances were evaluated using the F1 score, a harmonic mean of the recall and precision values that offers a compromise between the number of false positives (FP) and false negatives (FN) that is often utilized in deep learning (Sokolova and Lapalme, 2009).

### 2.4.1. Early fusion

The goal of early fusion is to create a new RGB image that preserves the colors and details from the VIS image while integrating the hot spots that dominate in the TIR image (Figure 4). To create this image, the VIS image was first decomposed into a YCbCr color space to calculate the luminance (Y), the blue-difference chroma (Cb) and the red-difference chroma (Cr) components (Sun et al., 2020). The Y represents the brightness in a digital image (Gopinathan and Gayathri, 2016) and is a component often used in data fusion (Chen et al., 2024; Yu et al., 2014). The Cb and Cr components carry the color information (Sun et al., 2020). Using the four bands (R, G, B, TIR), a principal component analysis (PCA) was performed, and the first component was selected (93% of the variance). The first component mainly takes the red and TIR band information with the following weights: 61% red, -4% green, -6% blue and 79% TIR. This component was used as the Y in the YCbCr color space. The inverse YCbCr was then applied to get the new RGB image: the fused image.

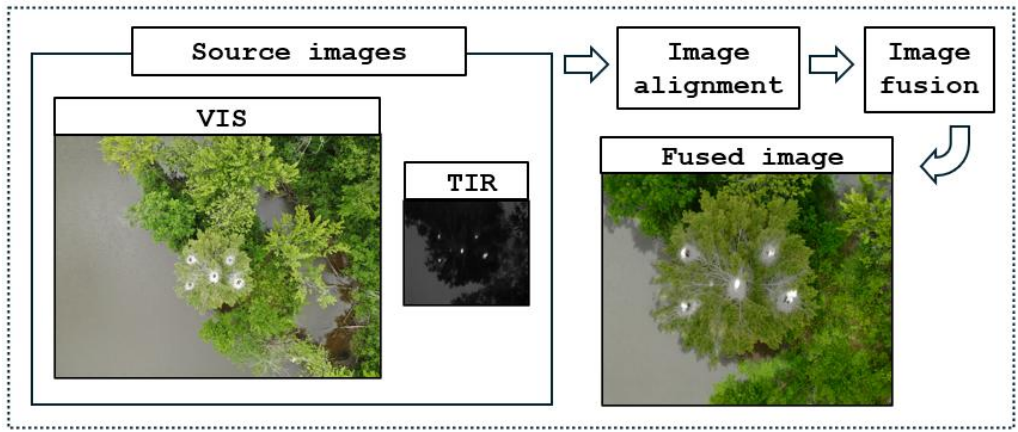

**Figure 4:** Early fusion method using corresponding nadir visible (VIS) and nadir thermal (TIR) images.

### 2.4.2. Late fusion

VIS images were used to train a model to detect three classes: occupied nest (nest with at least one individual inside), empty nest and isolated individual (herons outside a nest). TIR images were used to train a model to detect both isolated individuals and individuals inside nests in a single class "heron". The goal of the late fusion is to train a third classifier that will make the final decision based on the outputs from the individual VIS and TIR classifiers (Figure 5). The classes remain the same with the addition of the False Positive (FP) class.

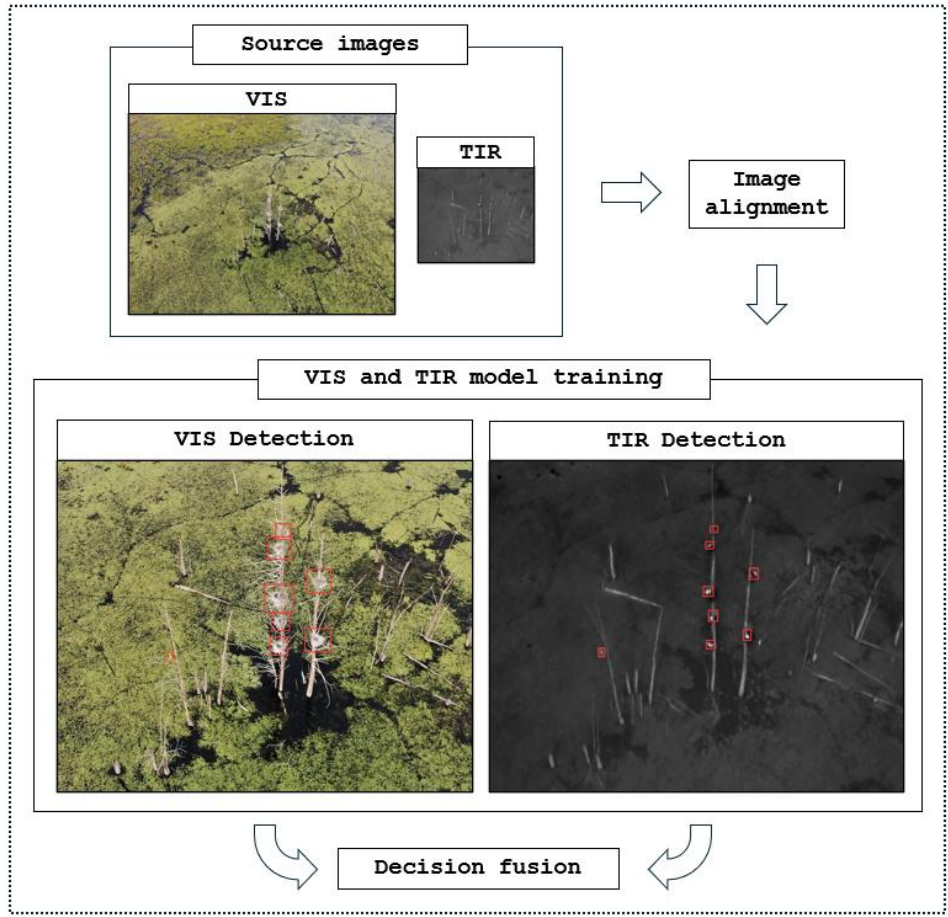

**Figure 5:** Late fusion process using visible (VIS) and thermal (TIR) detections (red boxes). Example using oblique corresponding VIS and TIR images.

The first step of the late fusion process is to geometrically intersect detection boxes from the VIS, the TIR and the ground truth (GT). We then used one hot encoding[1] to encode the VIS and the TIR decisions as well as the IoU (Intersection over Union) value, resulting in five features (occupied nest score, empty nest score, isolated individuals score, TIR score, IoU). Finally, a Classification and Regression Tree (CART) (Breiman et al., 1984) was trained to produce the final decision.

After the first stage, a fraction of the bounding boxes do not have any intersections, however, a decision still needs to be made, but only based on a single VIS or TIR modality. The non intersecting VIS decisions were kept unchanged and TIR decisions were classed as individual herons. Some of the boxes were found to be FP in the TIR and VIS images and so were classified as such by the late fusion classifier.

## 3. Results

### 3.1. Image alignment

The alignment and verification approach was able to rule out certain images that had too much alignment error and thus presented a risk of confusion during processing (Figure 6). After the alignment process, 39% of the available image pairs (images with a heron and/or a nest) were kept. The final dataset contained 755 image pairs that were separated into training (485), validation (121) and test (149) datasets while maintaining class representativeness (Table 2). For the VIS dataset, most of the labeled objects were occupied nests (87%). The empty nest and isolated individual classes represented 7% and 6% of the labeled objects, respectively. The TIR dataset had fewer labeled objects than the VIS dataset. These missing objects were mostly empty nests, which are almost never visible in TIR images. Most of the VIS and TIR image pairs were from forested heronries (70% oblique, 14% nadir). The rest of the VIS and TIR image pairs were acquired from aquatic heronries (15% oblique, 1% nadir).

---

[1] One hot encoding is a technique that converts categorical data into a binary format by creating a separate column for each class, where each row has the class score in the column representing its class and 0s elsewhere

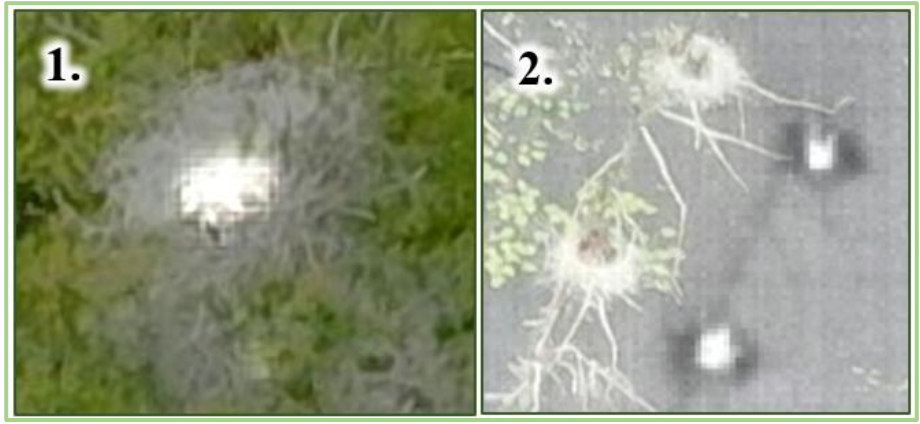

**Figure 6 :** Close-ups of great blue heron nests in early fusion images illustrating a good alignment (1), and an alignment error (2) between corresponding visible and thermal images.

**Table 2 :** Number of labeled objects in each dataset for the visible and thermal images.

| Class | Objects per dataset | | | |
| --- | --- | --- | --- | --- |
| | Training (485 images) | Validation (121 images) | Test (149 images) | Total (755 images) |
| **Visible images** | | | | |
| Occupied nest | 937 | 228 | 316 | 1481 (87%) |
| Empty nest | 74 | 17 | 35 | 126 (7%) |
| Isolated individual | 57 | 17 | 20 | 94 (6%) |
| Total | 1068 | 262 | 371 | 1701 |
| **Thermal images** | | | | |
| Heron (isolated individual and individuals inside nests) | 991 | 247 | 334 | 1572 |

### 3.2. Automated detection with YOLO11n

#### 3.2.1. Model training

The VIS and TIR models were trained on the same corresponding images to ensure comparability. Because the dataset is very unbalanced, so-called macro metrics (recall, precision and F1 score) were derived from the confusion matrix (Opitz and Burst, 2021) using the Scikit-learn library (Scikit-learn Developers, 2025). The F1 score was 85.4% for the TIR model and 72.2% for the VIS model (all classes) (Table 3). The occupied nest class represented 87% of the VIS dataset and had an F1 score of 87.9%. The F1 score was 66.0% for the empty nest class and 62.5% for the isolated individual class. This was to be expected, considering the low number of labeled objects in these classes.

**Table 3:** F1 score (0.5 threshold) on the validation set for thermal (TIR) and visible (VIS) models.

| Dataset | Classes | F1 score |
|---------|---------|----------|
| **TIR** | Heron | 85.4% |
| **VIS** | Occupied Nest | 87.9% |
|         | Empty Nest | 66.0% |
|         | Isolated individual | 62.5% |
|         | All classes | 72.2% |

### 3.2.2. Model comparison

The performance of the three approaches (VIS-only, early and late fusion) were compared on the same test set. The dataset was unbalanced, with occupied nests (Figure 7) constituting 85% of the total number of objects (316 objects), followed by empty nests (10%, 35 objects) and isolated individuals (5%, 20 objects).

The VIS-only model performed effectively, missing just 7% of ground truth detections and generating a low rate of FP (8%). However, the model's performance for the minority classes (empty nests and isolated individuals) was significantly lower compared to its strong performance on the majority class.

The early fusion model yielded fewer missed detections (4%) than the VIS-only model, indicating an improved ability to identify objects. Conversely, it resulted in a slightly higher FP rate (9%). Notably, the early fusion model demonstrated significant higher performance for the minority classes (empty nests and isolated individuals) compared to the VIS-only model. The early fusion model achieved an overall F1 score of 64.3%, which was the highest among all the models compared here.

As for late fusion, we observed a higher number of detections, as we not only kept intersecting bounding boxes between VIS and TIR images, but also boxes from single modalities (VIS or TIR-only). This higher number of detections can (at least partly) explain why the fraction of missed detections was the lowest compared to the previous two models, whereas the fraction of FPs was the highest among all the models.

Overall, both early and late fusion methods demonstrated added value compared to the VIS-only model for detecting GBHE. The early fusion model achieved the highest overall F1 score (Table 4) and showed better performance for the sparsely represented classes (empty nests and isolated individuals) (Table 5). Conversely, the late fusion model excelled in detecting the majority class (occupied nests) and recorded the fewest missed detections (Table 5). A key advantage of the late fusion approach is its less stringent requirement for pixel-level image alignment, unlike early fusion, which necessitates precise alignment and consequently discards a larger portion of image data.

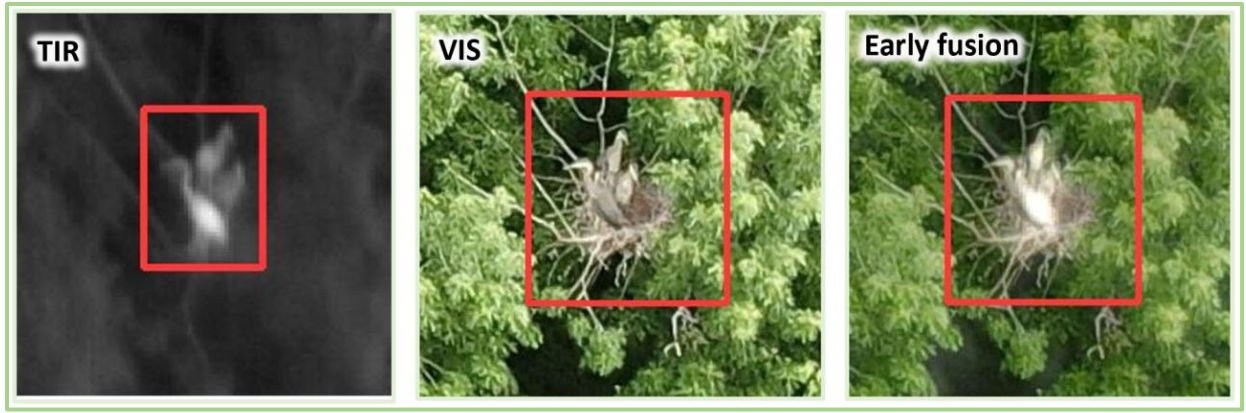

**Figure 7 :** Example of a correct detection in a corresponding thermal image (TIR), visible image (VIS) and early fusion image.

**Table 4:** Metrics computed from the confusion matrices on the test dataset for the visible (VIS), early fusion and late fusion models.

|  | Detections | False Negatives | False Positives | Average Recall | Average Precision | F1 score |
|---|---|---|---|---|---|---|
| **VIS-only** | 372 | 25 (7%) | 28 (8%) | 53.0% | 64.8% | 56.6% |
| **Early fusion** | 375 | 16 (4%) | 35 (9%) | 61.7% | 73.1% | 64.3% |
| **Late fusion** | 417 | 10 (3%) | 58 (14%) | 50.4% | 80.3% | 58.5% |

**Table 5 :** F1 score per class for the visible (VIS), early fusion and late fusion models on the test dataset.

|  | Occupied Nest | Empty Nest | Isolated individual |
|---|---|---|---|
| **VIS-only** | 90.2% | 44.0% | 35.7% |
| **Early fusion** | 88.8% | 49.1% | 55.6% |
| **Late fusion** | 93.0% | 44.4% | 38.1% |
| **Ground truth** | 316 (85%) | 35 (10%) | 20 (5%) |

## 4. Discussion

This study demonstrated the added value of using late fusion for detecting GBHE. To our knowledge, the results showed for the first time that the use of late fusion can be applied successfully on synchronous VIS and TIR aerial images to detect an animal species. The task of improving the detection results through early and late fusion was not easy, since they had to surpass the already well-performing VIS and TIR models. The CART algorithm gave better performances for the main class, occupied nest, while having less alignment restrictions than early fusion.

### 4.1. The challenges of image alignment

The automated image alignment process with LightGlue is a step forward compared to existing early-fusion studies applied to animal detection (for example, Krishnan et al. (2023), which performed manual steps to align images). The use of LightGlue proved to be an effective way of

https://doi.org/10.48550/arXiv.2511.22768

aligning the VIS and TIR images in challenging conditions. The automated process yielded enough aligned images pairs to test both early and late fusion. However, the alignment process discarded more than half of the available dataset, specifically when water was in the background, which caused a lack of corresponding features in both images. The early fusion method requires a more precise alignment, of one pixel or less, than late fusion, which uses the interception between the detection boxes. To compare the two methods, the same exact dataset was used. If only late fusion had been used, many more images would have been usable.

Further improvements to the alignment process could be made by using a more complex transformation to attempt to correct the optical offset between the two sensors. This step could augment the number of successfully aligned VIS and TIR images. Given that we used the pre-trained LightGlue model, some fine-tuning on our dataset could also be beneficial.

### 4.2. Early fusion: Combining VIS and TIR images at the input level

The early fusion method effectively compromises between VIS and TIR information, making GBHE hot spots visible while preserving the detailed VIS image background. This approach improved detection for minority classes (empty nests and isolated individuals) compared to the VIS-only model. However, the small sample size for these classes (35 empty nests, 20 isolated individuals) necessitates a cautious interpretation of these specific results. While early fusion reduced false negatives, it increased false positives compared to the VIS-only model. The primary limitation of early fusion remains its stringent requirement for precise pixel-level alignment between VIS and TIR images, a challenge discussed in Section 4.1.

Comparisons with existing literature show mixed results. Krishnan et al. (2023) reported that early fusion improved deer detection, attributing this to deer being a cryptic species often found in shadowed areas benefiting from combined VIS and TIR data. However, their study involved different YOLO models, had a smaller dataset, and lacked FP/FN reporting, limiting direct comparison. Sousa et al. (2023) found early fusion beneficial for human detection, primarily in low-light conditions where the VIS model performed poorly. In our study, where data did not present low-light issues, early fusion provided limited improvements for the main class and, in some cases, slightly lowered performance compared to the already well-performing VIS model. Our more complex early fusion method, which integrates PCA from four bands (RGB and TIR) into the YCbCr color space, appeared to yield better improvements compared to simpler four-band input methods described elsewhere.

### 4.3. Late fusion: Enhancing Detection Robustness through Decision-Level Integration

The late fusion method, which employs a CART algorithm, demonstrated enhanced performance for the occupied nest class compared to the VIS-only model. A key strength of this approach is the CART classifier's ability to efficiently identify false positives (FPs) with a 90% recall rate, as FPs were largely uncorrelated between the VIS and TIR sensors, thereby reducing decision-making confusion. Furthermore, the late fusion model successfully reduced the number of missed detections (false negatives) by effectively combining information from both modalities. For the minority classes (empty nests and isolated individuals), the late fusion model performed poorly, similar to the VIS-only model, which may be attributed to the limited features available for training the CART classifier (YOLO provides only the most probable class and its score per bounding box).

A significant advantage of late fusion is its reduced sensitivity to precise pixel-level image alignment compared to early fusion, allowing for the use of more image pairs even if detection boxes do not fully intersect. This flexibility broadens its applicability. Consistent with findings from Sousa et al. (2023), who observed late fusion improving human detection across various datasets (even those with well-performing individual VIS/TIR models), our study also found that late fusion improved results despite already proficient VIS and TIR baseline models. Other studies further support the efficacy of decision-level late fusion, noting its effectiveness for crowd counting, particularly at night (Cheng et al., 2024), and its superiority over early fusion for multi-defect detection on exterior walls (Yang et al., 2023). Our findings align with these trends, as late fusion surpassed early fusion for the primary class (occupied nest).

### 4.4. Limitations

The task of detecting GBHE nests was challenging, as nests were often partially or almost totally covered by vegetation in forested habitat, leaving only a partial target for the models. Isolated individuals were even more difficult to detect. Their smaller size compared to nests made them hard to even see during the labelling process. When only a small portion of their body is visible (i.e. the head), they can be barely recognisable. Therefore, it was not surprising that the isolated individual class did not perform well across all models. During the labelling process, the TIR image was useful to help locate herons in the VIS image when comparing the two. The TIR image was also helpful when deciding if a nest was occupied or not when analysing a VIS image. This complementary information can help decision making when the VIS image alone is not clear enough, and is essentially what happens in late fusion. Another element to consider here is the presence of great egret (*Ardea alba*) nests that the models sometimes confused with GBHE nests (Figure 8).

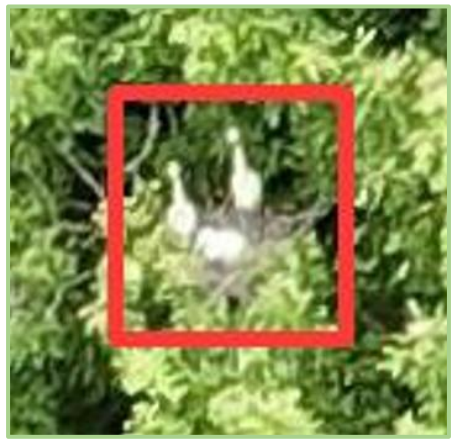

**Figure 8 :** Great egret (Ardea alba) mistakenly detected as an occupied nest (false positive) by the visible (VIS-only) model.

The relatively coarse ground sampling distance (GSD) of the VIS and TIR images was not ideal to locate hard to see objects. With a pixel representing a couple of centimetres, smaller objects like the head of an isolated individual were not very clear on the VIS and TIR images. With TIR images, any object hotter than the environment with a similar size as a GBHE has a high likelihood of being a FP, since the pixel size on the ground (4-5 cm) does not allow to clearly distinguish the shape of the GBHE. Although lowering the flight altitude could improve the GSD, it could also increase disturbance to the herons. Another way to improve the GSD would be to use a higher resolution VIS sensor, such as a DJI Zenmuse P1 with a GSD of about 1 cm/px at 50 m (Figure 9). Preliminary tests performed on very high-resolution VIS datasets (DJI Zenmuse P1 and DIJ Mavic 3 Enterprise) achieved an overall F1 score of 77.2% and class-specific F1 score of 93.6%, 69.7% and 68.2% for the occupied nest, empty nest and isolated individual classes, respectively, using YOLO11n. These results surpassed those presented in this study, especially for the last two classes.

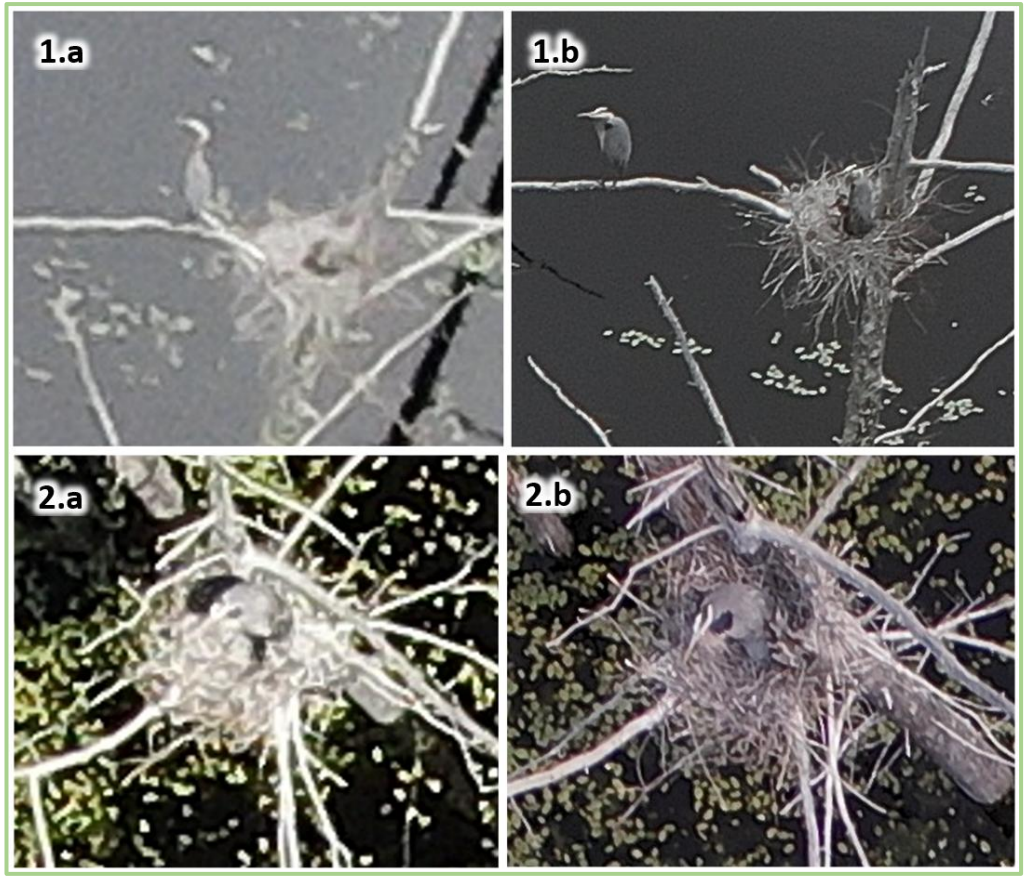

**Figure 9 :** Close-up of the great blue heron nest in oblique (1.) and nadir (2.) visible images from a DJI Zenmuse H20T sensor (a.) and DJI Zenmuse P1 sensor (b.). Both images taken from a DJI Matrice 300 at an oblique angle at 50 m altitude.

### 4.5. Implications for wildlife surveys

While early and late fusion models yielded better results than the VIS-only model, some aspects should be considered in the context of a survey. In GBHE surveys conducted in Quebec for example, the entirety of the heronry must be covered in order to properly count all the nests (Beaupré, 2021). Only surveying a portion (i.e., a sampling) of the heronry would not get an accurate count of all the nests, as they do not have a homogenous distribution throughout the environment. The use of early and late fusion models requires VIS and TIR image alignment, which can discard a large portion of the dataset (61% in this case study). In a real survey, this would mean neglecting a large portion of the heronry. Moreover, from an operational point of view, the small FOV of the fused image implies longer drone flights to cover the heronry when compared to the wide FOV from a VIS sensor. In this context, the use of VIS-only model would be optimal, considering the relatively low added value of the fusion models. However, the added value of the TIR images could be more beneficial for a more cryptic species than the GBHE, such as the white-tailed deer (Krishnan et al., 2023) or for crepuscular or nocturnal species. Since GBHE nests are usually located on the top of trees, they are rarely in the shade. The added information from the TIR image can help locate the animal when it is blending in its environment or in low light conditions.

This study further proves that the use of a Deep Learning approach to automatically detect an animal species on aerial images is effective. Since a manual count on images would be tedious and very time-consuming, automated detection enables this task to be carried out considerably faster (Delplanque et al., 2023a). In recent years, the use of CNNs for animal detection was tested on several species with different CNN architectures (Delplanque et al., 2023b; Eikelboom et al., 2019; Noguchi et al., 2025; Peng et al., 2020). While detection of wildlife has repeatedly been proven to be highly effective with CNNs, there is still a need for an accurate counting method on multi-angle image datasets. Further research is needed to achieve an accurate count of GBHE in an operational context. One avenue worth exploring would be a camera-mounted aircraft, since the heronry surveys in Quebec are already conducted by helicopters. The data can be post-processed, making the survey process faster as real-time visual counting would not be necessary. An aircraft equipped with very high-resolution sensors could allow the presented method to be used, while alleviating the drone's very limited coverage area. Moreover, the use of video recording could possibly improve surveys by allowing the tracking and counting of animals (Gonzalez et al., 2016).

**Acknowledgements**

This project was supported by the Applied Research and Development grants from the Natural Sciences and Engineering Research Council of Canada (NSERC) (CCARD-2022-00270). Profound thanks are offered to all involved in the data acquisition process: Sébastien Auger, Francis Bédard, Sophiane Béland, Virginie Boivin, Yan Bourque, Charles-Antoine Brassard, Émilie Chalifour, Chantale Côté, Alexandre Lajeunesse, Christian Pilon, Véronique St-Hilaire, Guillaume Tremblay, Émilie Trépanier, members of the Animal Care Committee from MELCCFP, as well as Batistin Bour and Mélanie Dussault from CERFO, Émilie Dorion and Hugues Tennier from Parc National du Mont Tremblant, Pourvoirie Roger Gladu, Louise Dumont and finally, several landowners for access to heronries.

**Author's contribution**

Jérôme Lemaitre and Mathieu Varin conceived the initial project ideas; Camille Dionne-Pierre, Samuel Foucher and Jérôme Théau designed the methodology and analysed the data; Camille Dionne-Pierre, Patrick Charbonneau and Maxime Brousseau collected the data; Camille Dionne-Pierre led the writing of the manuscript with critical inputs from Samuel Foucher and Jérôme Théau; all authors contributed to the drafts and gave approval for publication.

**Author's Declaration**

All authors have reviewed and approved the final version of the manuscript. Each author has made significant contributions to the work, and all individuals meeting the criteria for authorship have been included.